\def\year{2021}\relax
\documentclass[letterpaper]{article} 
\usepackage{aaai21}  
\usepackage{times}  
\usepackage{helvet} 
\usepackage{courier}  
\usepackage[hyphens]{url}  
\usepackage{graphicx} 
\usepackage{natbib}  
\usepackage{caption} 
\usepackage{amsmath}
\usepackage{array}
\usepackage{caption}
\usepackage{subcaption}
\usepackage{url}
\usepackage[multiple]{footmisc}

\nocopyright
\usepackage[switch]{lineno}  %

\title{Reinforcement Learning for Optimization of COVID-19 Mitigation Policies}
\iftrue
\author {
        Varun Kompella\textsuperscript{*} \textsuperscript{\rm 1},
        Roberto Capobianco\textsuperscript{*} \textsuperscript{\rm 1, \rm 2},
        Stacy Jong\textsuperscript{\rm 3},
        Jonathan Browne\textsuperscript{\rm 3},
        Spencer Fox\textsuperscript{\rm 3},
        Lauren Meyers\textsuperscript{\rm 3},
        Peter Wurman\textsuperscript{\rm 1},
        Peter Stone\textsuperscript{\rm 1, \rm 3}\\
}
\affiliations {
    \textsuperscript{\rm 1} Sony AI \\
    \textsuperscript{\rm 2} Sapienza University of Rome \\
    \textsuperscript{\rm 3} The University of Texas at Austin \\
    \textsuperscript{*} Joint First Authors, varun.kompella@sony.com, roberto.capobianco@sony.com \\
    
}
\else
\author{\ }
\fi

\iffalse
\long\def\commentp#1{{\bf --PS: #1--}}
\long\def\commentpw#1{{\bf --PW: #1--}}
\long\def\commentsam#1{{\bf --Sam: #1--}}
\long\def\commentcraig#1{{\bf --Craig: #1--}}
\long\def\commenttom#1{{\bf --Tom: #1--}}
\else
\long\def\commentp#1{}
\long\def\commentpw#1{}
\long\def\commentsam#1{}
\long\def\commentcraig#1{}
\long\def\commenttom#1{}
\fi

\long\def\oursim{\textsc{PandemicSimulator}}

\begin{document}
\maketitle

\newcommand{\covid}{\textsc{covid-19}}

\begin{abstract}
The year 2020 has seen the \covid\ virus lead to one of the worst global pandemics in history. As a result, governments around the world are faced with the challenge of protecting public health, while 
keeping the economy running to the greatest extent possible. Epidemiological models provide insight into the spread of these types of diseases and predict the effects of possible intervention policies. 
However, to date, the even the most data-driven intervention policies rely on heuristics.  \commenttom{seems argumentative, I would say something like even the most data-driven intervention policies rely on heuristics -- FIXED} In this paper, we study how reinforcement learning (RL) can be used to optimize mitigation policies that minimize the economic impact without overwhelming the hospital capacity. 
\commentsam{Minor: It feels a little weird to say you introduce the RL method without saying you introduce the sim first, but maybe it's fine -- FIXED by PRW}
Our main contributions are (1) a novel agent-based pandemic simulator which, unlike traditional models, is able to model fine-grained interactions among people at specific locations in a community; 
\commentcraig{Drop novel. If it wasn't novel it probably wouldn't be a contribution. -- KEEP PRW}
and (2) an RL-based methodology for optimizing fine-grained mitigation policies within this simulator. Our results validate both the overall simulator behavior and the learned policies under realistic conditions.
\end{abstract}

\section{Introduction}

Motivated by the devastating \covid\ pandemic, much of the scientific community, across numerous disciplines,
\commentcraig{multiple $\rightarrow$ numerous  -- FIXED PRW}
is currently focused on developing safe, quick, and effective methods to prevent the spread of biological viruses, or to otherwise mitigate the harm they cause.  These methods include vaccines, treatments, public policy measures, economic stimuli, and hygiene education campaigns.  Governments around the world are now faced with high-stakes decisions regarding which measures to enact at which times, often involving trade-offs between public health and economic resiliency. 
When making these decisions, governments often rely on epidemiological models that predict and project the course of the pandemic. \commentcraig{Is there a difference between predict and project in your usage?  -- KEEP PRW, there is a difference.}

The premise of this paper is that the challenge of mitigating the spread of a pandemic while maximizing personal freedom and economic activity is fundamentally a sequential decision-making problem: the measures enacted on one day affect the challenges to be addressed on future days.  As such, modern reinforcement learning (RL) algorithms 
are well-suited to optimize government responses to pandemics. 

For such learned policies to be relevant, they must be \commenttom{For such policies to be relevant, they must be learned learned using -- FIXED} trained within an epidemiological model that accurately simulates the 
spread of the pandemic, as well as the effects of government measures.  To the best of our knowledge, none of the existing epidemiological simulations have the resolution to allow reinforcement learning to explore the regulations that governments are currently struggling with.
\commentsam{I don't feel like I care that they're not developed to support RL. I'd emphasize the things they lack: the ability to simulate at a pretty granular scale I believe.  -- FIXED PRW}
\commentcraig{I agree with Sam. What don't the current models let us do? Why do we need a better approach?}

Motivated by this, our main contributions are:
\begin{enumerate}
\item The introduction of \oursim, a novel \commentcraig{novel :( -- KEEP PRW} open-source\footnote{\url{https://github.com/SonyAI/PandemicSimulator}} agent-based simulator that models the interactions between individuals at specific locations within a community.  Developed in collaboration between AI researchers and epidemiologists (the co-authors of this paper), \oursim\ models realistic effects such as testing with false positive/negative 
rates, imperfect public adherence to social distancing measures, contact tracing, and variable spread rates among infected individuals.  
Crucially, \oursim\ models community interactions at a level of detail that allows the spread of the disease to be an emergent property of people's behaviors and the government's policies.
\commentsam{I don't know if I like talking about it being designed for RL here as much as saying that it allows granular changes to policy, which sure enables RL, but it enables other techniques too. -- FIXED PRW} \commentsam{Coming back from further down, I guess other agent-based sims exist, so it's probably hard to separate out what the RL-appropriate features are here, but I still don't love this}  An interface with OpenAI Gym~\cite{brockman2016openai} is provided to enable support for standard RL libraries;
\item A demonstration that a reinforcement learning algorithm can indeed identify a policy that outperforms a range of reasonable baselines within this simulator; \commenttom{Give a number.  Outperforms by how much?-- WONT FIX, too hard PRW}
\item An analysis of the resulting learned policy, which may provide insights regarding the relative efficacy of past and potential future \covid\ mitigation policies.
\end{enumerate}
While the resulting policies have \emph{not} been implemented in any real-world communities, this paper establishes the potential power of RL in an agent-based simulator, and may serve as an important first step towards real-world adoption.

The remainder of the paper is organized as follows. We first discuss related work \commentcraig{If it wasn't related to this paper I don't think it would be 'related'. -- FIXED PRW} and then introduce our simulator in Section \ref{sec:simulator}. Section \ref{sec:reop_policies} presents our reinforcement learning setup, while results are reported in Section \ref{sec:experiments}. Finally, Section \ref{sec:conclusions} reports some conclusions and directions for future work. \commentsam{nit: I don't like the wording of "future work ideas", maybe "directions for future work" -- FIXED RC}
\commenttom{If you need space you can drop this paragraph - not really needed in a conf submission}

\section{Related Work}

 
Epidemiological models differ based on the level of granularity in which they track individuals and their disease states. ``Compartmental'' models group individuals of similar disease states together, assume all individuals within a specific compartment to be homogeneous, and only track the flow of individuals between compartments~\cite{Tolles2020}. While relatively simplistic, these models have been used for decades and continue to be useful for both retrospective studies and forecasts as were seen during the emergence of recent diseases \cite{Rivers2020,Metcalf2020,Cobey2020}. \commentcraig{A `recent emerging disease' is not an event---during does not apply. 'with recent'? `during recent outbreaks'? `during the emergence of recent diseases'? -- FIXED RC}


However, the commonly used macroscopic (or mass-action) compartmental models are not appropriate when outcomes depend on the characteristics of heterogeneous individuals. In such cases, network models \cite{bansal2007individual,Liu12680,Khadilkar2020} and agent-based models \cite{grefenstette2013fred, del2013modeling,aleta2020modelling} may be more useful predictors. \commentsam{Better suited for what, I don't feel like "better suited" stands on its own -- FIXED PRW} Network models \commentsam{"in fact" doesn't add anything and feels stilted in my opinion} encode the relationships between individuals as static connections in a contact graph along which the disease can propagate.
Conversely, agent-based simulations, \commentsam{If you have the space to say "agent based simulations" instead of "the latter" here, I think it would be easier to read -- FIXED RC} such as the one introduced in this paper, explicitly track individuals, their current disease states, and their interactions with other agents over time. Agent-based models allow one \commentcraig{missing word. `us'? `one'? -- FIXED RC} to model as much complexity as desired---even to the level of simulating individual people and locations as we do---and thus enable one to model people's interactions at offices, stores, schools, etc.  Because of their increased detail, they enable one to study the hyper-local interventions that governments consider when setting policy. For instance, \citet{larremore2020test} simulate the SARS-CoV-2 dynamics both through a fully-mixed mass-action model and an agent-based model representing the population and contact structure of New York City.
 
\oursim\ has the level of details needed to allow us to apply RL to 
optimize dynamic government intervention policies (sometimes referred to as
``trigger analysis'' 
e.g. \citealt{Duque2020.04.29.20085134}).  RL has been
applied previously to several mass-action models \cite{libin2020deep,song2020reinforced}. 
These models, however, do not take into account individual behaviors or any complex interaction patterns. 
The work that is most closely related to our own includes both the SARS-CoV-2 epidemic simulators from \citet{Hoertel2020} and \citet{aleta2020modelling}, which model individuals grouped into households who visit and interact in the community. While their approach builds accurate contact networks of real populations, it doesn't allow us to model how the contact network would change as the government intervenes. 
\commentsam{What about it doesn't supoprt RL? Maybe you don't have the space for it, but I feel like that's useful info -- FIXED PRW} \citet{xiao2020modeling} construct a detailed, pedestrian level simulation that simulates transmission indoors and study three types of interventions. 
\citet{liu2020microscopic} presents a microscopic approach to model epidemics, which can explicitly consider the consequences of individuals' decisions on the spread of the disease. Multi-agent RL is then used to let individual agents learn to avoid infections. 
\commentcraig{This paragraph comes off negative (it has the kind of feel I was always got in trouble for as a grad student). Instead can you simply describe what they did do and then say what your model allows us to do differently? -- FIXED PRW}

For any model to be accepted by real-world decision-makers, they must
be provided with a reason to trust that it accurately models the
population and spread dynamics in their own community.  For both
mass-action and agent-based models, this trust is typically best
instilled via a model calibration process that ensures that the model
accurately tracks past data.  For example, \citet{Hoertel2020} perform a calibration using daily mortality data until 15 April. 
Similarly, \citet{libin2020deep} calibrate their model based on the symptomatic cases reported by the British Health Protection Agency for the 2009 influenza pandemic. \citet{aleta2020modelling}, instead, only calibrate the weights of intra-layer links by means of a rescaling factor, such that the mean number of daily effective contacts in that layer matches mean number of daily effective contacts in the corresponding social setting. 
While not a main focus of our research, we have taken initial steps to demonstrate
that our model can be calibrated to track real-world data, as
described in Section~\ref{sec:calibration}.

\section{PandemicSimulator: A COVID-19 Simulator}
\label{sec:simulator}

\begin{figure}[t]
  \centering
  \includegraphics[width=0.9\columnwidth]{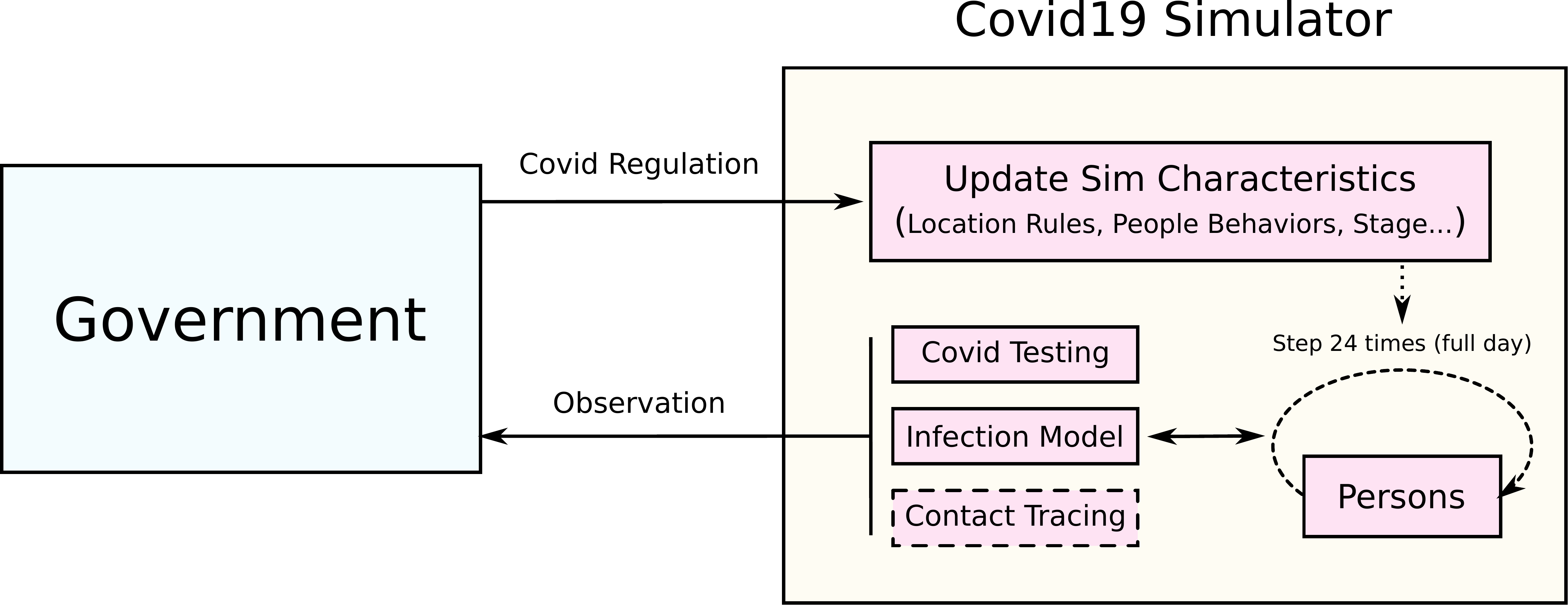}
  \caption{Block diagram of the simulator.}
  \label{fig:sim_descr}
\end{figure}

The functional blocks of \oursim, shown in \figurename~\ref{fig:sim_descr}, are:
\begin{itemize}
\item \emph{locations}, with properties that define how people interact within them;
\item \emph{people}, who travel from one location
to another according to individual daily schedules;
\item an \emph{infection model} that updates the infection state of each
person;
\item an optional \emph{testing strategy} that imperfectly exposes the infection state of the
population;
\item an optional \emph{contact tracing} strategy that identifies an 
infected person's recent contacts;
\item a \emph{government} that makes policy decisions.
\end{itemize}

\commentsam{Why are you putting the figure at the top via [t]? It feels like it's the wrong place in my opinion -- SHOULD BE FIXED NOW WHEN REMOVING COMMENTS RC}

The simulator models a day as 24 discrete hours, with each person potentially changing \commentsam{nit: I'd say "potentially changing locations each hour", it just reads better to me -- FIXED RC}
locations each hour. At the end of a day, each person's infection state is updated.
The government interacts with the environment by declaring
\emph{regulations}, which impose restrictions on the people and locations.
If the government activates testing, the simulator identifies a set of people to 
be tested and (imperfectly) reports their infection state. If contact tracing is
active, each person's contacts from the previous days are updated. \commentsam{is this also imperfect? - WONTFIX because it is currently perfect PRW}
The updated perceived infection state and other state variables are returned as 
an observation to the government. The process iterates as long as the infection 
remains active.
\commentsam{I don't like one sentence paragraphs, so I'd jam this into the previous one -- FIXED RC} The following subsections describe the functional 
blocks of the simulator in greater detail.\footnote{We relegate some implementation details to an appendix. 
}

\subsection{Locations}

Each location has a set of attributes that specify when the location is open, 
what roles people play there (e.g. worker or visitor), and the maximum number of people of each role. These
attributes can be adjusted by regulations, such as when the government
determines that businesses should operate at half capacity.
Non-essential locations can be completely closed by the government.
The location types used in our experiments are \emph{homes}, \emph{hospitals}, 
\emph{schools}, \emph{grocery stores}, \emph{retail stores}, and \emph{hair salons}. 
The simulator provides interfaces to make it easy to add new location types.

One of the advantages of an agent-based approach is that we can more 
accurately model variations in the way people interact
in different types of locations based on their roles. The base location class
supports workers and visitors, and defines a \emph{contact rate}, $b^{\text{loc}}$, as
a 3-tuple $(x, y, z) \in [0, 1]^3$, where $x$ is the worker-worker rate, $y$ is the 
worker-visitor rate, and $z$ is the visitor-visitor rate.
These rates are used to sample interactions every hour in each
location to compute disease transmissions. For example,
consider a location that has a contact rate of $(0.5, 0.3, 0.4)$ and 10 workers
and 20 visitors. \commentsam{Move "In expectation," to the start -- FIXED RC} In expectation, a worker would make contact with
5 co-workers and 6 visitors in the given hour. 
Similarly, \commentsam{comma after similarly -- FIXED RC} a visitor would be expected to make contact with 3 workers and 8 other
visitors. 
Refer to our supplementary material (Appendix~\ref{sec:parameters}, Table~\ref{table:sim_params})
for a listing of the contact rates 
and other parameters for all location types used in our experiments.

The base location type can be extended for more complex situations. 
For example, a \emph{hospital} adds an additional role (critically sick patients), a 
capacity representing ICU beds, and contact rates between workers and patients.  


\subsection{Population}

A \emph{person} in the simulator is an automaton that has a state and
a person-specific behavior routine. These routines create
person-to-person interactions throughout the simulated day
and induce 
dynamic contact networks.

Individuals are assigned an age, drawn from the distribution of 
the US age demographics, and are randomly assigned to be either high risk or of normal health. \commentcraig{What is this classification based on? -- FIXED PRW}
Based on their age, each person is categorized as either a
\emph{minor}, a \emph{working adult} or a \emph{retiree}. Working adults are 
assigned to a work location, and minors to a school, which they attend 8 hours
a day, five days a week. Adults and retirees are assigned favorite hair salons
which they visit once a month, and grocery and retail stores which they visit once a week. \commentsam{Are they assigned a set of retail stores that they reliably visit every week? That feels odd. I'd have expected there to be a set of retail stores that they visit with some probability. Also, fixing it to be a week feels odd here, I'd have figured that you'd sample the times between visits from distribution. It's a little messier this way, but probably too late to change that :)  - WONT FIX PRW}
Each person has a compliance parameter that determines the
probability that the person flouts regulations each hour. 
\commenttom{I disagree with Sam here.  Having stared at actual grocery foot traffic data for the last 6 months, I can tell you there is extremely strong weekly periodicity.}

The simulator constructs households from this population such that 
$15\%$ house only retirees, and the rest have at least one working adult 
and are filled by randomly assigning the remaining children, adults, and
retirees. To simulate informal social interactions, households may attend social events
twice a month, subject to limits on gathering sizes. \commentsam{I'd have expected to draw the number from a distribution, where some households meet frequently and some never do -- WONT FIX PRW }

At the end of each simulated day, the person's infection state is updated
through a stochastic model based on all of that individual's interactions 
during the day (see next section). Unless otherwise prescribed 
by the government, when a person becomes ill they follow their routine. 
However, even the most basic government interventions require sick people
to stay home, and at-risk individuals to avoid large gatherings. If a person becomes critically ill, they are 
admitted to the hospital, assuming it has not reached capacity.


\subsection{SEIR Infection Model}
\label{sec:seir}

\commentsam{Why is the figure on the top here too? Putting it before the section always feels weird to me. Bottom feels like it would be better again. -- SHOULD BE FIXED RC}
\commentsam{The legend characters are real small here. Maybe that's just how it needs to be, but it's really hard to see the characters in the legend's boxes. -- Should be fixed VK}
\oursim\ implements a modified SEIR (susceptible, exposed, infected,
recovered) infection model, as shown in \figurename~\ref{fig:seir}. 
See supplemental Appendix~\ref{sec:parameters}, Table~\ref{table:seir_params} for specific parameter values and the transition probabilities of the SEIR model.  Once exposed to the virus, an individual's
path through the disease is governed by the transition probabilities. However, 
the transition from the susceptible state ($S$)
to the exposed state $(E)$ requires a more detailed explanation.

At the beginning of the simulation, a small, randomly selected set of individuals 
seeds the pandemic in the latent non-infectious, exposed state ($E$). The rest of the population
starts in $S$. The exposed individuals soon transition
to one of the infectious states and start interacting with susceptible people.
For each susceptible person $i$, the probability they become infected on a given day,
$P^{S \rightarrow E}_{i}(\textit{day})$, is calculated based on their
contacts with infectious people that day.
\begin{align} 
\label{eq:infection-probability}
&P^{S \rightarrow E}_{i}(\textit{day}) = 1 - \prod_{t=0}^{23} \overline{P}^{S \rightarrow E}_{i}(t)
\end{align}
\noindent
\commenttom{If there is a citation to others that use a similar formula that would help the case here -- WONT FIX PRW}
where $\overline{P}^{S \rightarrow E}_{i}(t)$ is the
probability that person $i$ is \textit{not} infected at hour $t$. \commentsam{Would it be simpler to use the notation $P^{S \rightarrow S}_i(t)$? It feels simpler than the bar version to me. -- I'D KEEP RC}
Whether a susceptible person becomes infected in a given hour depends on whom 
they come in contact with. Let $\mathcal{C}^{j}_{i}(t) = \{\textit{p} \overset{b^{j}}{\sim}
N_{j}(t)|\textit{p} \in N^{\text{inf}}_{j}(t)\}$ be the set of
infected contacts of person $i$ in location $j$ at hour $t$ where
$N^{\text{inf}}_{j}(t)$ is the set of infected persons in location $j$
at time $t$, $N_{j}(t)$ is the set of all persons in $j$ at time $t$, and
$b^{j}$ is a hand-set contact rate for $j$. \commentsam{I'd say that $b^j$ is the location contact rate from the previous section -- NOT SURE THIS IS WRONG? RC} To model the variations
in how easily individuals spread the disease, each individual $k$ has an 
infection spread rate, $a^{k} \sim \mathcal{N}^{\text{bounded}} (a, \sigma)$ sampled from a
bounded Gaussian distribution. Accordingly,  
\begin{align} 
  \label{eq:hourly-not-infection-probability}
  &\overline{P}^{S \rightarrow E}_{i}(t) = \prod_{k \in
  \mathcal{C}^{j}_{i}(t)} (1 -
    a^{k}).
\end{align}
\begin{figure}[t]
  \centering
  \includegraphics[width=0.9\columnwidth]{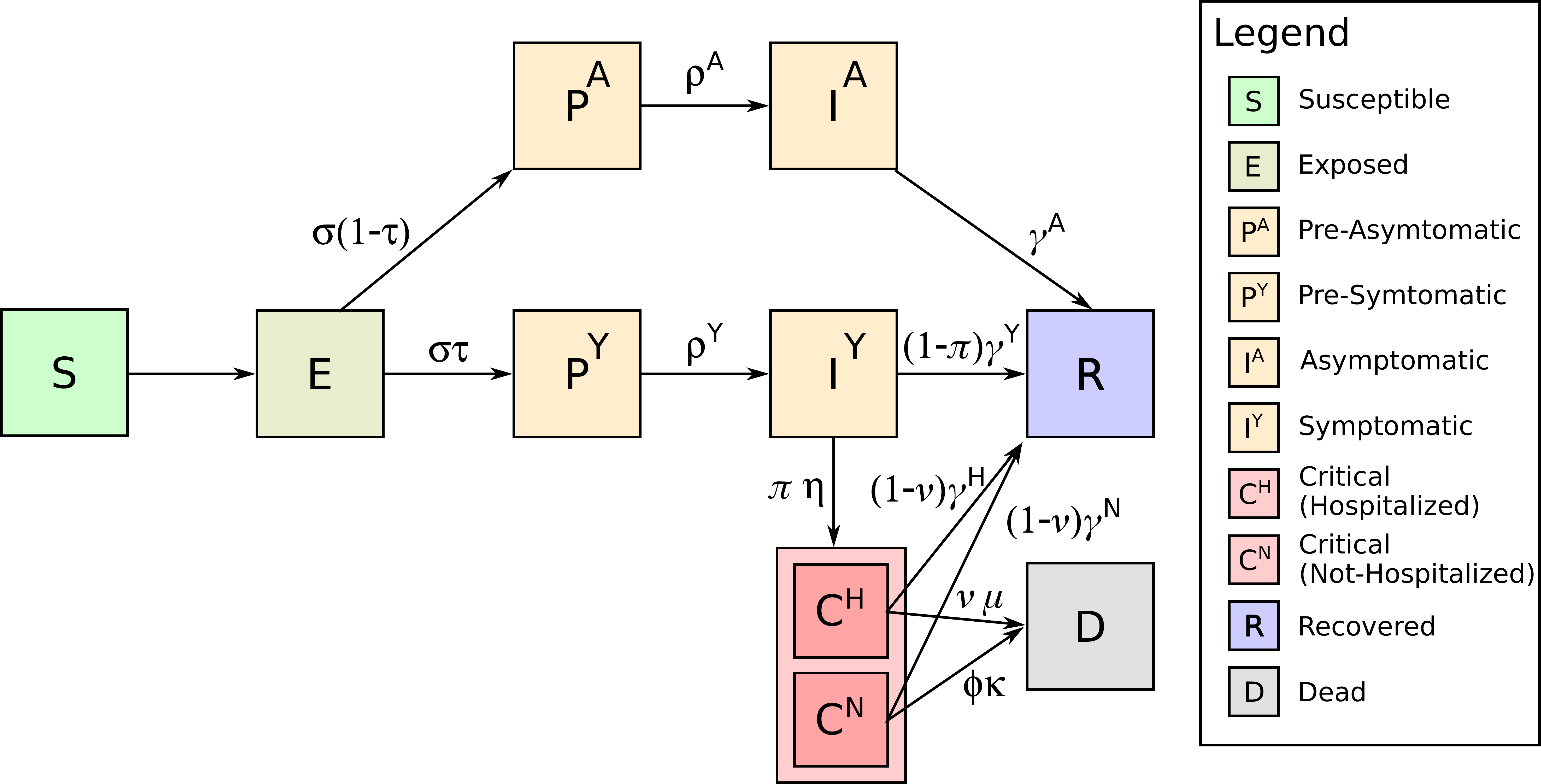}
  \caption{SEIR model used in \oursim}
  \label{fig:seir}
\end{figure}
\commentcraig{So your model assumes that a recovered person cannot be reinfected in the span of the experiment? If so you should make a comment to that end, because that's one of the big questions right now - and an assumption that Sweden built their approach off of I believe.-- PRW WONT FIX unless there is room.}

\subsection{Testing and Contact Tracing}
\label{sec:testing}

\oursim\ features a testing procedure to identify positive cases
of \covid. We do not model concomitant illnesses, so every critically sick or dead
person is assumed to have tested positive. Non-symptomatic and symptomatic 
individuals---and individuals that previously tested positive---get tested all at different configurable rates. Additionally, we model false positive and false negative test results.
Refer to the supplementary material (Appendix~\ref{sec:parameters}, Table~\ref{table:sim_params})
for a listing of the testing rates used in our experiment.

The government can also implement a contact tracing strategy 
that tracks, over the last $N$ days, the number of times each
pair of individuals interacted. When activated, this procedure allows the
government to test or quarantine all recent 1\textsuperscript{st}-order contacts and their 
households when an individual tests positive for \covid.

\subsection{Government Regulations}
\label{sec:obs_action}

As discussed earlier (see \figurename~\ref{fig:sim_descr}), the government
announces regulations to try to control the pandemic. 
The government can impose the following rules:
\begin{itemize}
\item social distancing: a value $\beta \in [0, 1]$ that scales the contact
  rates of each location by $(1 - \beta)$. 0 corresponds to
  unrestricted interactions; 1 eliminates all interactions;
\item stay home if sick: a boolean.  When set, people who have tested positive are requested to stay at home;
\item practice good hygiene: a boolean.  When set, people are requested to practice
  better-than-usual hygiene. 
\item wear facial coverings: a boolean.  When set, people are instructed to wear
  facial coverings. 
\item avoid gatherings: a value that indicates the maximum recommended size of gatherings. These values can differ
for high risk individuals and those of normal health;
\item closed businesses: A list of non-essential business
  location types that are not permitted to open.
\end{itemize}
\commentsam{I'd add a reminder that people don't always follow the rules here -- WONT FIX here PRW}


These types of regulations, modeled after government policies 
seen throughout the world, are often bundled into progressive \emph{stages} 
to make them easier to communicate to the population. Refer to Appendix~\ref{sec:parameters}, Tables~\ref{table:sim_params}-\ref{table:five_stage_regulation} for details on the parameters, their sources and the values set for each stage.

\subsection{Calibration}
\label{sec:calibration}
\oursim\ includes many parameters whose values are still poorly known, such as the spread rate of \covid\ in grocery stores
and the degree to which face masks reduce transmission. 
We therefore consider these parameters as free variables that can be used to \emph{calibrate} the simulator to match
the historical data that has been observed around the world.  
These parameters can also be used to 
customize the simulator to match a specific community.  A discussion
of our calibration process and the values we chose to model \covid\ are discussed
in Appendix~\ref{sec:app-calibration}. \commentcraig{Hmmm, I was waiting to see this information, but you tossed it in the appendix :( - WONT FIX, no room PRW}

\section{RL for Optimization of Regulations}
\label{sec:reop_policies}

\commentsam{Instead of "minimize the impact" maybe "minimize the spread", because "impact" could mean many things as you immediately point out -- FIXED RC}
An ideal solution to minimize the spread of a new disease like \covid\
is to eliminate all non-essential interactions and quarantine infected people
until the last infected person has recovered. However, the window to
execute this policy with minimal economic impact is very small. Once
the disease spreads widely this policy becomes impractical and the potential 
negative impact on the economy becomes enormous.
\commentsam{A better wording? "In practice around the world, we have seen ..." I definitely wouldn't start this sentence with "What" -- FIXED RC}
In practice,
around the world we have seen a strict \commentcraig{"harsh" carries implications of judgement. How about 'strict'? -- FIXED RC} lockdown followed by a gradual reopening that attempts
to minimize the growth of the infection while allowing partial economic activity.
Because \covid\ is highly contagious, has a long incubation period, and large 
portions of the infected population are asymptomatic, managing the reopening without
overwhelming healthcare resources is challenging. In this section, we tackle this sequential decision making problem using reinforcement learning (RL;~\citealt{sutton2018reinforcement}) to optimize the reopening policy.
\commenttom{Simpler rewrite: In this section, we tackle this sequential decision making problem using reinforcement learning to optimize the reopening policy -- FIXED PRW}

To define an RL problem we need to specify the environment, observations, actions, and rewards. \commentcraig{Ok, you've defined all the different components of an RL problem, but haven't described the RL problem, i.e., learning a policy which maximizes the sum of future rewards. If I didn't already know RL I don't think I would be clear what was going to happen or what you were optimizing. -- FIXED below PRW}

\subsubsection{Environment:} The agent-based pandemic simulator \oursim\ is the environment.\footnote{For the purpose of our experiments,  
we assume no vaccine is on the horizon and
that survival rates remain constant.  In practice, one may want to model the effect of improving survival rates as the medical community gains experience treating the virus.} 
\commenttom{You can drop this implementation detail - it was mentioned earlier. -- FIXED PRW}

\subsubsection{Actions:} The government is the learning agent. Its goal is to maximize its reward over the horizon of the pandemic.
Its action set is constrained to a pool of escalating stages, which it can either 
increase, decrease, or keep the same when it takes an action. Refer to Appendix~\ref{sec:parameters}, Table~\ref{table:five_stage_regulation} for detailed descriptions of the stages. 
\commentsam{Maybe similar to the "for the sake of realism" below, we should remind readers that people aren't forced to follow the stages, but maybe that's overkill - WONT FIX PRW}

\subsubsection{Observations:} At the end of each simulated day,
the government observes the environment. For the sake of realism, the
infection status of the population is partially observable, accessible only via statistics 
reflecting aggregate (noisy) test results and number of hospitalizations.\footnote{The simulator tracks ground truth data, like the number of people in each infection state, for evaluation and reporting.}

\subsubsection{Rewards:} We designed our reward function to
encourage the agent to keep the number of persons in critical 
condition ($n^c$) below the hospital's capacity ($C^{\text{max}}$),
while keeping the economy as unrestricted as possible. To this end, we use a reward that is a weighted sum of two objectives:
\begin{align}
  r = a~\max\left(\frac{n^c-C^{\text{max}}}{C^{\text{max}}},~0 \right) +
  b~\frac{\text{stage}^{p}}{\max_{j} \text{stage}^{p}_{j}}
  \label{eqn:reward}
\end{align}
where $\text{stage} \in [0, 4]$ denotes one of the 5 stages with $\text{stage}_4$ being the most restrictive. $a$, $b$ and $p$ are set to $-0.4$, $-0.1$ and $1.5$, respectively, in our experiments.
To discourage frequently changing restrictions, we also use a small shaping reward (with $-0.02$ coefficient) proportional to $|stage(t-1) - stage(t)|$. This linear mapping of stages into a $[0,1]$ reward space is arbitrary; if \oursim\ were being used to make real policy decisions, policy makers would use values that represent the real economic costs of the different stages.
\commentsam{I'd apparently mention that this the choice of reward functions taken here, but policy makers could give input to the tradeoffs in general. -- FIXED PRW}
\commentcraig{I think I must have missed something here. How is 'stage' defined numerically? Should the following be subscripted by t: r, n, stage (in the numeratore)? -- Fixed VK}

\subsubsection{Training:} We use the discrete-action Soft Actor Critic 
(SAC; ~\citealt{Haarnoja2018}) off-policy RL algorithm to optimize a 
reopening policy, where the actor and critic networks are two-layer 
deep multi-layer perceptrons with 128 hidden units.
One motivation behind using SAC over deep Q-learning approaches such as 
DQN~\cite{mnih2015human} is that we can provide the true infection summary 
as inputs to the critic while letting the actor see only the 
observed infection summary. Training is episodic with each episode 
lasting 120 simulated days. At the end of each episode, the 
environment is reset to an initial state. 
Refer to \commenttom{to -- FIXED} Appendix~\ref{sec:parameters}, Table~\ref{table:sim_params} for learning parameters.

\section{Experiments}
\label{sec:experiments}
\begin{figure}
  \centering
  \includegraphics[width=\columnwidth]{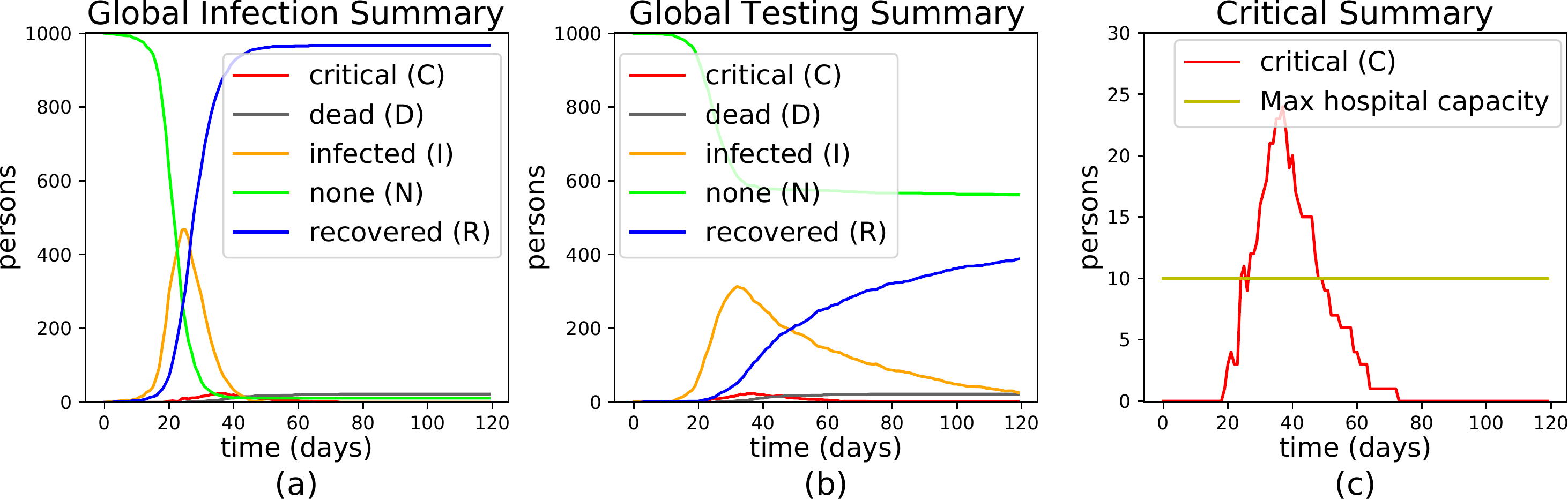}
  \caption{A single run of the simulator with no government restrictions, showing (a)
    the true global infection summary (b) the perceived infection state, and (c)
    the number of people in critical condition over time. \commentsam{Figure 3 is probably placed incorrectly, might just be in the commented version at this point though}}
  \label{fig:single_run_stage_0}
\end{figure}

\begin{figure*}[ht!]
  \centering
  \includegraphics[width=0.9\textwidth]{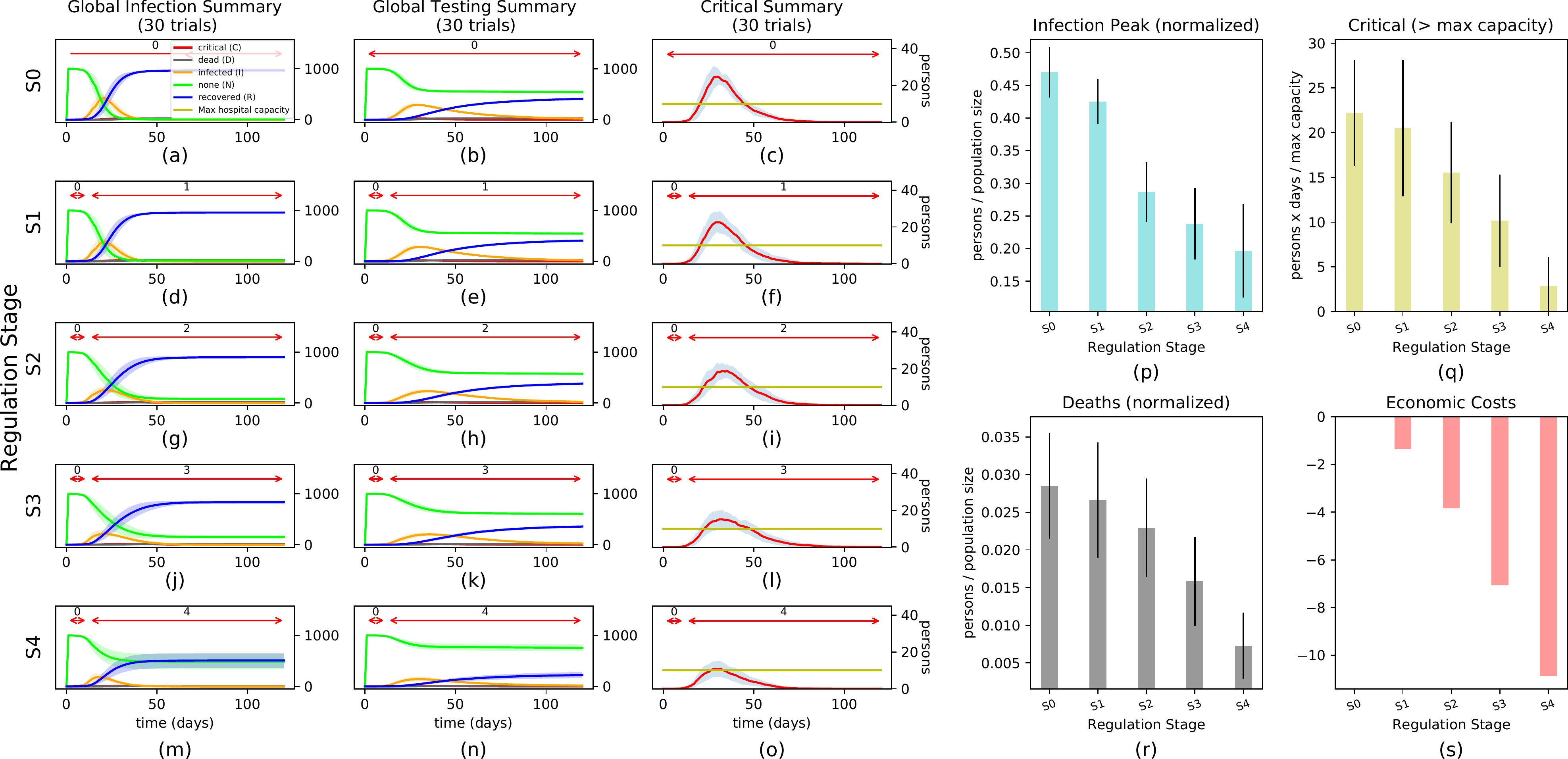}
  \caption{Simulator dynamics at different regulation stages. The
    plots are generated based on 30 different randomly seeded runs of
    the simulator. Mean is shown by a solid line and variance either
    by a shaded region or an error line. In the left set of graphs, the red line at the top indicates what regulation stage is in effect on any given day. \commentsam{The legend in this figure isn't legible, can we blow it up and have it on the side?. Or did you mean to remove it entirely like the second and third columns?} \commentsam{I don't think that the arrows are helping here since you're not switching stages.} \commentsam{Does dying not have an economic cost? Wait, why is the y axis days? -- Fixed VK}}
  \label{fig:staged_experiment}
\end{figure*}

The purpose of \oursim\ is to enable a more realistic evaluation of potential government policies for pandemic mitigation.
In this section, we validate that the simulation behaves as expected under controlled conditions, illustrate some of the many analyses it facilitates, and most importantly, demonstrate that it enables optimization via RL.

Unless otherwise specified, we consider a community size of 1,000 \commentsam{comma in 1,000 -- FIXED RC. NOT FIXED: I dislike the use of 1k and 10k in the footnote and 1,000 here, but I don't know if it's worth changing} \commentcraig{I'd like some statement addressing the small population size. Was this a computational limitation?} and a hospital
capacity of 10.\footnote{\oursim\ can easily handle larger experiments at the cost of greater time and computation.  Informal experiments showed that results from a population of 1k are generally consistent with results from a larger population when all other settings are the same (or proportional). Refer to Table~\ref{table:execution_times} in the appendix for simulation times for 1k and 10k population environments.}  
To enable calibration with real data, we limit government actions to five regulation stages similar to those
used by real-world cities\footnote{Such as at~\url{https://tinyurl.com/y3pjthyz}} (see
appendix for details), and assume the government does not act until at least five
people are infected. 

\figurename~\ref{fig:single_run_stage_0} shows plots of a single
simulation run with no government regulations
(Stage 0). \figurename~\ref{fig:single_run_stage_0}(a) shows the number of
people in each infection category per day. Without
government intervention, all individuals get infected, with the infection
peaking around the $25^{\text{th}}$
day. \figurename~\ref{fig:single_run_stage_0}(b) shows the metrics observed
by the government through the lens of testing and hospitalizations. 
This plot illustrates how the government sees information that is
both an underestimate of the penetration and delayed in time from the true state. Finally, \figurename~\ref{fig:single_run_stage_0}(c) 
shows that the number of people in critical condition goes
well above the maximum hospital capacity (denoted with a yellow line) resulting in
many people being more likely to die.
The goal of a good reopening policy is to keep the red curve
below the yellow line, while keeping as many businesses open as possible.

\figurename~\ref{fig:staged_experiment} shows plots of our infection
metrics averaged over 30 randomly seeded runs. Each row in
{\figurename}s~\ref{fig:staged_experiment}(a-o) shows the results of
executing a different (constant) regulation stage (after a short initial S0 phase), where S4 is the most restrictive and S0
is no restrictions. \commentcraig{This should be explained sooner. It affects the interpretation of the reward. --Fixed VK.} As expected, 
{\figurename}s~\ref{fig:staged_experiment}(p-r) show that the infection peaks, critical cases
and number of deaths are all lower for more restrictive stages. One way of explaining the effects of these regulations is that the government restrictions alter the connectivity of the contact graph. For example, in the experiments above, under stage 4 restrictions there are many more connected components in the resulting contact graph than in any of the other 4 cases.  See Appendix~\ref{sec:supplements} for details of this analysis.

Higher stage restrictions,
however, have increased socio-economic costs (\figurename~\ref{fig:staged_experiment}(s); computed using the second objective in Eq.\ \ref{eqn:reward}).
\commentsam{Does dying not have any cost? I would think that it does, but I guess you're not modeling it. I feel like you should at least acknowledge it somewhere though, but maybe I missed it. -- WONT FIX Good point, but too hard to fix PRW}
Our RL experiments illustrate how these competing objectives can be
balanced.

A key benefit of \oursim's agent-based approach is that it enables us to evaluate more dynamic policies\footnote{In this paper, \commentsam{comma here -- FIXED RC} we use the word ``policy'' to mean a function from state of the world to the regulatory action taken.  It represents both the government's policy for combating the pandemic (even if heuristic) and the output of an RL optimization.} than those described above.
In the remainder of this section we compare a set of
hand constructed policies, examine (approximations) of two real country's
policies, and study the impact of contact tracing. 
In Appendix~\ref{sec:sensitivity-analisys} we also provide an analysis of the model's sensitivity to its parameters. 
Finally, we demonstrate the application of RL to construct dynamic polices
that achieve the goal of avoiding exceeding hospital capacity while minimizing
economic costs. 
As in \figurename~\ref{fig:staged_experiment}, 
throughout this section we report our results using plots that are
generated by executing 30 simulator runs with fixed seeds. 
All our experiments were run on a single core, using an Intel
i7-7700K CPU $@$ 4.2GHz with 32GB of RAM.

\subsection{Benchmark Policies}



\begin{figure*}
  \centering
  \includegraphics[width=0.9\textwidth]{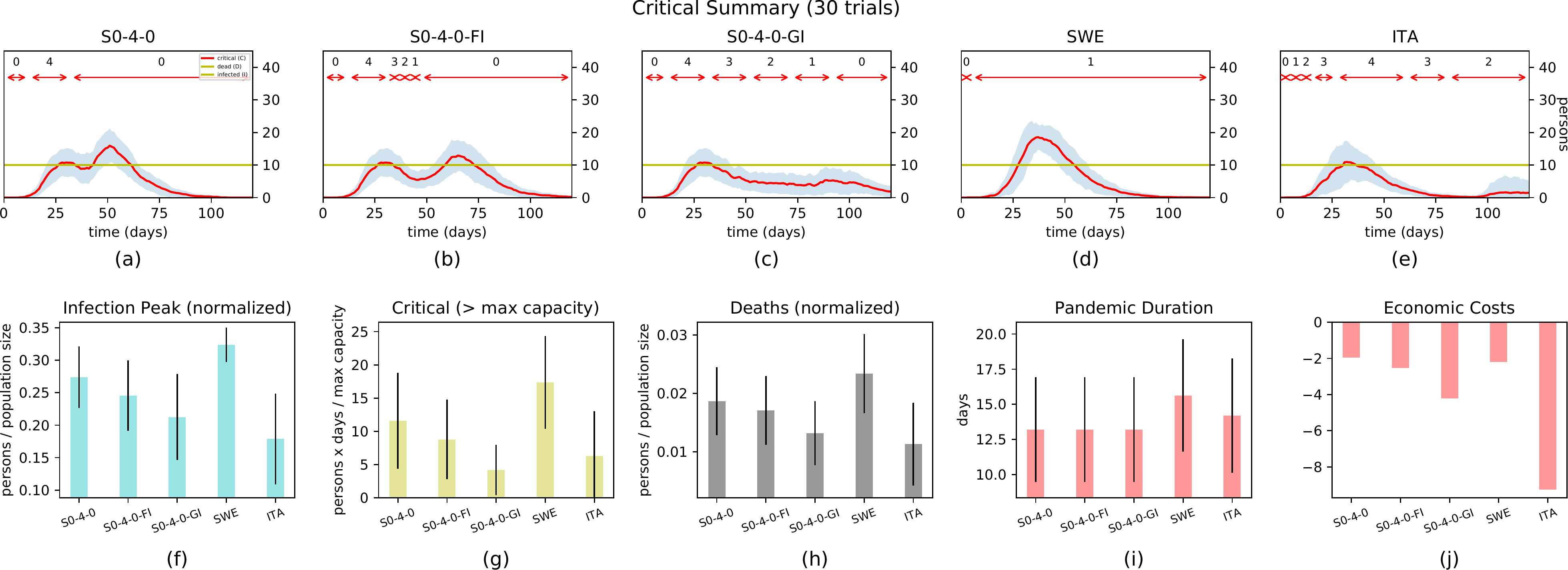}
  \caption{Simulator dynamics under different hand constructed and reference government policies.}
  \label{fig:handset_government_policies}
\end{figure*}

\begin{figure*}[t]
  \centering
  \includegraphics[width=0.9\textwidth]{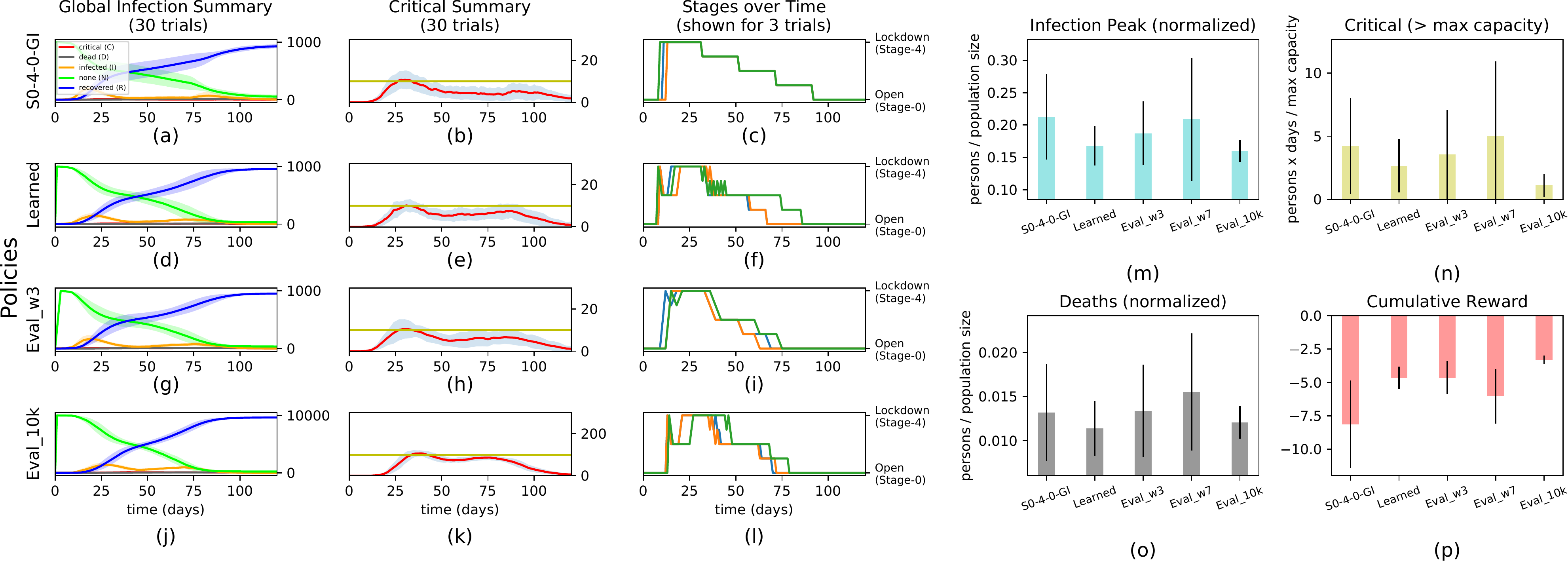}
  \caption{Simulator runs comparing the S0-4-0-GI heuristic policy with a learned policy. The figure also shows results of the learned policy evaluated at different action frequencies and in a larger population environment.}
  \label{fig:policies}
\end{figure*}

To serve as benchmarks, we defined three heuristic and two policies inspired by real governments'
approaches to managing the pandemic.
\begin{itemize}
    \item \textbf{S0-4-0}: Using this policy, the government switches from stage
      0 to 4 after reaching a threshold of 10 infected
      persons. After 30 days, it switches directly back to stage 0;
      
    \item \textbf{S0-4-0-FI}: The government starts like
      S0-4-0, but after 30 days it executes a fast, incremental (FI) return
      to stage 0, with intermediate stages lasting 5 days;
      
    \item \textbf{S0-4-0-GI}: This policy implements a more gradual incremental (GI)\commentsam{Maybe "gradual incremental (GI)" -- FIXED RC}
      return to stage 0, with each intermediate stage lasting 10 days;
      
    \item \textbf{SWE}: This policy represents the one adopted by the Swedish 
  government, which recommended, but did not require remote work, and was generally unrestrictive.\footnote{\url{https://tinyurl.com/y57yq2x7}; \url{https://tinyurl.com/y34egdeg}} 
  Table~\ref{table:swedish_regulation} in the supplementary material shows how we mapped this policy into a 2-stage action space. 

    \item \textbf{ITA}: this policy represents the one adopted by the
  Italian government, which was generally much more restrictive.\footnote{\url{https://tinyurl.com/y3cepy3m}}
  Table~\ref{table:italian_regulation} shows our mapping of this policy to a 5-stage action space.
\end{itemize}

\figurename~\ref{fig:handset_government_policies} compares the
hand constructed policies. From the point of view of minimizing overall
mortality, S0-4-0-GI performed best. In particular, slower re-openings ensure longer
but smaller peaks. While this approach leads to a second wave right
after stage 0 is reached, the gradual policy prevents hospital capacity from
being exceeded.

\figurename~\ref{fig:handset_government_policies} also contrasts the approximations of the policies employed by Sweden and Italy in the early stages of the pandemic (through February 2020). 
The ITA policy leads to fewer
deaths and only a marginally longer duration. However, this simple 
comparison does not account for the economic cost of policies, 
an important factor that is considered by decision-makers.

\subsection{Testing and Contact Tracing}
\label{sec:testing-contact-tracing-res}
To validate \oursim's ability to model testing and contact tracing we compare several strategies with different testing rates and contact horizons. We consider daily testing rates of 0.02, 0.3, and 1.0 (where 1.0 represents the extreme case of everyone being tested every day) and contact tracing histories of 0, 2, 5, or 10 days.  For each condition, we ran the experiments with the same 30 random seeds. The full results appear in Appendix~\ref{sec:app_testingcontact_tracing}.

Not surprisingly, contact tracing is most beneficial with higher testing rates and longer contact histories because more testing finds more infected people and the contact tracing is able to encourage more of that person's contacts to stay home. Of course, the best strategy is to test every person every day and quarantine anyone who tests positive. Unfortunately, this strategy is impractical except in the most isolated communities. Although this aggressive strategy often stamps out the disease, the false-negative test results sometimes allow the infection to simmer below the surface and spread very slowly through the population.

\subsection{Optimizing Reopening using RL}
\label{sec:rl}

A major design goal of \oursim\ is to support optimization of re-opening policies using RL.  In this section, we test our hypothesis that a learned policy can outperform the benchmark policies.
%
Specifically, RL optimizes a policy that (a) is adaptive to the changing infection state, (b) keeps the number of critical patients below the hospital threshold, and (c) minimizes the economic cost.

We ran experiments using the 5-stage regulations defined in Table~\ref{table:five_stage_regulation} (Appendix~\ref{sec:parameters});  trained the policy by running RL optimization for roughly 1 million training steps; 
and evaluated the learned policies across 30 randomly seeded initial conditions. 
Figures~\ref{fig:policies}(a-f) show results comparing our best heuristic policy (S0-4-0-GI) 
to the learned policy. 
The learned policy is better across all metrics as shown in Figures~\ref{fig:policies}(m-p). Further, we can see how the learned policy reacts to the state of the pandemic; Figure~\ref{fig:policies}(f) shows different traces through the regulation space for 3 of the trials. The learned policy briefly oscillates between Stages 3 and 4 around day $40$. To minimize such oscillations, we evaluated the policy at an action frequency of one action every 3 days (bi-weekly; labeled as Eval\_w3) and every 7 days (weekly; labeled as Eval\_w7). Figure~\ref{fig:policies}(p) shows that the bi-weekly variant \commenttom{variant? - FIXED} performs well, while making changes only once a week \commenttom{approach? -- FIXED} slightly reduces the reward.  To test robustness to scaling, we also evaluated the learned policy (with daily actions) in a town with a population of 10,000 (Eval\_10k) and found that the results transfer well. This success \commenttom{success? -- FIXED} hints at the possibility of learning policies quickly even when intending to transfer them to large cities. 


This section presented results on applying RL to optimize reopening policies.  An interesting next step would be to study and explain the learned policies as simpler rule based strategies to make it easier for policy makers to implement. For example, in Figure~\ref{fig:policies}(l), we see that the RL policy waits at stage 2 before reopening schools to keep the second wave of infections under control. Whether this behavior is specific to school reopening is one of many interesting questions that this type of simulator allows us to investigate.


\section{Conclusion}
\label{sec:conclusions}

Epidemiological models aim at providing predictions regarding the effects of various possible 
intervention policies that are typically manually selected. In this paper, instead, we introduce 
a reinforcement learning methodology for optimizing  adaptive  mitigation  policies  aimed  at  maximizing the degree to which  the  economy  can  remain  open  without  overwhelming the local 
hospital capacity. To this end, we implement an open-source agent-based simulator, where pandemics 
can be generated as the result of the contacts and interactions between individual agents in a 
community. We analyze the sensitivity of the simulator to some of its main parameters and
illustrate its main features, while also showing that adaptive policies optimized via RL achieve better 
performance when compared to heuristic policies and policies representative of those used in the real world.

While our work opens up the possibility to use machine learning to explore fine-grained policies 
in this context, \oursim\ could be expanded and improved in several directions. One important direction for 
future work is to perform a more complete and detailed calibration of its parameters against
real-world data. It would also be useful to implement and analyze \commenttom{analyze-- FIXED} additional testing and contact 
tracing strategies to contain the spread of pandemics.

\section*{Ethics Statement}

This paper is intended as a proof of concept that Reinforcement Learning algorithms have the potential to optimize government policies in the real world.  We acknowledge that the question of what policies to enact is a highly polarizing issue with many political and socio-economic implications.  As described in detail in the paper, the simulator introduced here has many free parameters that can dramatically affect its behavior.  While we have made an effort to calibrate it to some real-world data, this effort was mainly for the purpose of showing that the simulator \emph{can} be calibrated.  If it is to be used to inform any real world policy decisions, it will be essential for these parameters to be calibrated to match historical data in the community in question, in conjunction with local experts.  Similarly, the available government actions would need to be set according to the options available to local policy-makers.  Even so, it would be important to recognize that the simulator encodes several assumptions and is inherently approximate in its projections.  Policymakers must be fully informed of these assumptions and limitations before they draw any conclusions or take any actions based on our experiments or any future experiments in \oursim.

These cautionary considerations notwithstanding, we consider the contributions of this paper to be an important first step towards the prospect of optimizing pandemic response policies via RL.  We would like nothing more than for this work to be continued (by us or by others) to the point where it can be used to good effect for the purpose of saving lives and/or improving the economic health in real world communities.

\bibliography{aaai21.bib}

\clearpage

\appendix

\section{Supplementary Material}
\label{sec:supplements}

\subsection{Simulation Parameters}
\label{sec:parameters}

In Section \ref{sec:simulator} of the paper, we introduced \oursim\, which is a very flexible tool with which we study the propagation of the disease and the effects of various government regulations on that propagation. As such, it also has a lot of parameters that control its behaviour. These parameters, many of which we introduced in the main paper, are loosely grouped into three categories.
\begin{itemize}
    \item Environmental: control the size of the population, the number and types of locations, and the ways people interact in those locations (Table~\ref{table:sim_params}).
    \item Epidemiological: control the progression of the disease in an individual (Table~\ref{table:seir_params}). These can be changed to model different pandemics.
    \item Regulatory: control the impacts and efficacy of the government regulations (Tables~\ref{table:five_stage_regulation}, ~\ref{table:swedish_regulation}, ~\ref{table:italian_regulation}). Regulations also determine the size of the action space for reinforcement learning experiments.
\end{itemize}
The rest of this appendix details the parameter settings used in this experiments in the paper.

\begin{table*}[h]
  \footnotesize
  \caption{Environment Parameters (vetted by epidemiologists) used in \oursim}
  \label{table:sim_params}
  \renewcommand{\arraystretch}{1.7} 
  \begin{tabular}{| m{12em} | m{20em} | m{20em} |}
    \hline
    \textbf{Parameter} & \textbf{Value} & \textbf{Source}\\
    \hline
    age &  Based on us population age distribution & \url{https://www.populationpyramid.net/united-states-of-america/2018/}\\
    \hline
    
    1k population parameters \newline (number, worker capacity, visitor capacity) & 
        Homes: (300, -, -) \newline
        Grocery stores: (4, 5, 30) \newline
        Offices: (5, 200, 0) \newline
        Schools: (1, 40, 300) \newline
        Hospitals: (1, 30, 0) \& patient capacity of 10\newline
        Retail stores: (4, 5, 30)\newline
        Hair salons: (4, 3, 5)\newline
        Cemetery: (1, -, -)
    & Hospital capacity is set based on the prescriptions from the
French Red Cross on hospital building, which suggest to
have between 8 and 11 hospital beds every 1000 people. Rest of the values are
based on our best guess reflecting a small colony.\\
    \hline
    
    10k population parameters \newline (number, worker capacity, visitor capacity) & 
        Homes: (3000, -, -) \newline
        Grocery stores: (10, 10, 30) \newline
        Offices: (50, 200, 0) \newline
        Schools: (3, 40, 300) \newline
        Hospitals: (1, 80, 0) \& patient capacity of 100\newline
        Retail stores: (15, 10, 30)\newline
        Hair salons: (20, 3, 5)\newline
        Cemetery: (1, -, -)
    & 1k population hospital capacity is scaled by 10 and also to match with 2020 data from the American Hospital Association (924,107 total beds distributed
among 6,146 hospitals in the US $\approx$ 150 beds per hospital on an average). Rest of the values are
based on our best guess reflecting a small town.\\
    \hline
    
    Infection spread rates\newline
    (mean, standard deviation) & (0.03, 0.01) & Loosely calibrated to fit Sweden's \covid~ hospitalizations and deaths.\\
    \hline
    
    Location contact rates \newline $(b^{\text{loc}}, b^{\text{loc}}_{\text{min}})$ &
        Homes: (0.5, 0.3, 0.3), (0, 1, 0) \newline
        Grocery stores: (0.2, 0.25, 0.3), (0, 1, 0) \newline
        Offices: (0.1, 0.01, 0.01), (2, 1, 0) \newline
        Schools: (0.1, 0., 0.1), (5, 1, 0) \newline
        Hospitals: (0.1, 0., 0.), (0, 3, 1) \newline
        Retail stores: (0.2, 0.25, 0.3), (0, 1, 0)\newline
        Hair salons: (0.5, 0.3, 0.1), (1, 1, 0)\newline
        Cemetery: (0., 0., 0.05), (0, 0, 0)
        & Based on our best guess.\\
    \hline
    
    Testing parameters & 
        Random testing rate: 0.02 \newline
        Symptomatic testing rate: 0.3 \newline
        Critical testing rate: 1.0 \newline
        False positive rate: 0.001 \newline
        False negative rate: 0.01 \newline
        Re-test previous-positive rate: 0.033
        & Based on our best guess. \\
    \hline
    
    Wear facial coverings 
    \newline (spread rate multiplier) & 0.6 & \url{https://www.lhsfna.org/index.cfm/lifelines/may-2020/how-effective-are-masks-and-other-facial-coverings-at-stopping-coronavirus/}\\
    \hline
        
    Practice good hygiene
    \newline (spread rate multiplier) & 0.8 &  Based on our best guess.  \\
    \hline
        
    Rule compliance hour probability & 0.99 & \\
    \hline
    
    Social gatherings & 
    House parties (duration of 5 hours) once every month in each house on random dates. & All house parties are open-invite events. This is done to represent all other gatherings like concerts, sporting events, etc.\\
    \hline
    
    RL critic inputs & 
    Global infection summary, stage & Critic is only used during training.\\
    \hline
    
    RL actor inputs & 
    Global testing summary, stage & To keep it realistic.\\
    \hline
    
    RL Actions &
    [-1, 0, 1] & Stage change\\
    \hline
    
    Critic and actor networks & 1 hidden layer of 128  ReLU units each &\\
    \hline
    
    Simulator steps per action & 24 & A new action at the start of each day \\
    \hline
    
    Learning rates &
    Critic: 1e-3, actor: 1e-4 & \\
    \hline 
    
    SAC entropy coefficient $\alpha$ & 0.01 & \\
    \hline
    
    Stale network refresh rate & 0.005 &\\
    \hline
    
    RL discount factor & 0.99&\\
    \hline

  \end{tabular}
\end{table*}

\begin{table*}[h]
  \footnotesize
  \caption{Epidemiological Parameters used in our SEIR model. Values given as
    five-element vectors are age-stratified with values corresponding
    to 0-4, 5-17, 18-49, 50-64, 65+ year age groups, respectively.}
  \label{table:seir_params}
  \renewcommand{\arraystretch}{1.7} 
  \begin{tabular}{| m{18em} | m{20em} | m{14em} |}
    \hline
    \textbf{Parameter} & \textbf{Value} & \textbf{Source}\\
    \hline
    $\sigma$: exposed rate & $\frac{1}{\sigma}\sim Tr(1.9,2.9,3.9)$ & \citet{Zhang2020}\\
    \hline
    $\tau$: symptomatic proportion (\%) &
                                          $57$  & \citet{Gudbjartsson2020} \\
    \hline
    $\rho^Y$: pre-symptomatic rate &
                                     $\frac{1}{\rho^Y}=2.3$  & \citet{He2020}\\
    \hline
    $\rho^A$: pre-asymptomatic rate &
                                      $\frac{1}{\rho^A}=2.3$  & \\
    \hline
    $\gamma^Y$: recovery rate in symptomatic non-treated compartment &
                                                                       $\frac{1}{\gamma^Y}\sim Tr(3.0,4.0,5.0)$  & \citet{He2020}\\
    \hline
    $\gamma^A$: recovery rate in asymptomatic compartment &
                                                            $\frac{1}{\gamma^A}\sim Tr(3.0,4.0,5.0)$  & \\
    \hline
    $\gamma^H$: recovery rate in hospitalized compartment &
                                                            $\frac{1}{\gamma^H}\sim Tr(9.4,10.7,12.8)$  & Fit to Austin admissions \& discharge data
                                                                                                          (Avg=10.96. 95\% CI = 9.37 to 12.76)\\
    \hline
    $\gamma^N$: recovery rate in hospitalization needed compartment &
                                                            $0.0214$  & \\
    \hline
    $\text{YHR}$: symptomatic case hospitalization rate (\%), age and
    risk specific
 &
   Overall:  $[0.07018,  0.07018,  4.735, 16.33, 25.54]$,
   Low risk: $[ 0.04021,  0.03091,  1.903,  4.114,  4.879]$,
   High risk: $[ 0.4021,  0.3091, 19.03, 41.14, 48.79]$

                                        & Adjusted from \citet{Verity2020}\\
    \hline
    $\text{HFR}$: hospitalized fatality ratio, age specific (\%) &
                                                                   $[4, 12.365, 3.122,
                                                                   10.745,
                                                                   23.158]$
                                        & Computed from the infected
                                          fatality ratio in \citet{Verity2020}\\
    \hline
    $\pi$: rate of symptomatic individuals go to hospital &
                                                      $\pi=\frac{\gamma^Y\text{YHR}}{\eta
                                                            +
                                                            (\gamma^Y-\eta)\text{YHR}}$
                                        & \\
    \hline
    $\eta$: rate from symptom onset to hospitalized &
                                                      $0.1695$ & \citet{Tindale2020}\\
    \hline
    $\mu$: rate from hospitalized to death &
                                             $\frac{1}{\mu}\sim Tr(5.2,8.1,10.1)$ & Fit to Austin admissions \& discharge data
                                                                                    (Avg=7.8, 95\% CI = 5.21 to 10.09)\\
    \hline
    $\nu$: death rate on hospitalized individuals &
                                                    $\nu = \frac{\gamma^H\text{HFR}}{\mu+(\gamma^H-\mu)\text{HFR}}$ &

    \\
    \hline
    $\phi$: death rate on individuals that need hospitalization &
                                                    $[0.239, 0.3208, 0.2304, 0.3049, 0.4269]$ &

    \\
    \hline
    $\kappa$: rate from hospitalization needed to death &
                                             $0.3$ &\\
    \hline
  \end{tabular}
\end{table*}

\begin{table*}[h]
\footnotesize
  \caption{Five stage Covid regulations}
  \label{table:five_stage_regulation}
  \renewcommand{\arraystretch}{1.7} 
  \begin{tabular}{| m{4em} | m{10em} | m{8em} | m{8em} | m{9em} | m{8em} |}
    \hline
    \textbf{Stages} & \textbf{Stay home if sick, \newline Practice good hygiene} & \textbf{Wear facial \newline coverings} & \textbf{Social \newline distancing} & \textbf{Avoid gathering size \newline (Risk: number)} & \textbf{Locked locations} \\
    \hline
    Stage 0 & False & False & None & None & None\\
    \hline
    Stage 1 & True & False & None & Low: 50, High: 25 & None\\
    \hline
    Stage 2 & True & True & 0.3 & Low: 25, High: 10  & School, Hair Salon\\
    \hline
    Stage 3 & True & True & 0.5 & Low: 0, High: 0 & School, Hair Salon\\
    \hline
    Stage 4 & True & True & 0.7 & Low: 0, High: 0 & School, Hair Salon, Office, Retail Store\\
    \hline
  \end{tabular}
\end{table*}

\begin{table*}[h]
\footnotesize
  \caption{Swedish Covid regulations. Note that, while the Swedish
  government recommended, but did not require, remote work and had different 
  recommendations for different ages of school children, we mapped the
  overall policy to be roughly stage 1 reported here.}
  \label{table:swedish_regulation}
  \renewcommand{\arraystretch}{1.7} 
  \begin{tabular}{| m{4em} | m{10em} | m{10em} | m{8em} | m{10em} | m{8em} |}
    \hline
    \textbf{Stages} & \textbf{Stay home if sick, \newline Practice good hygiene} & \textbf{Wear facial coverings} & \textbf{Social distancing} & \textbf{Avoid gathering size \newline (Risk: number)} & \textbf{Locked locations} \\
    \hline
    Stage 0 & False & False & None & None & None\\
    \hline
    Stage 1 & True & False & 0.7 & Low: 50, High: 50 & None\\
    \hline
  \end{tabular}
\end{table*}

\begin{table*}[h]
\footnotesize
  \caption{Italian Covid regulations}
  \label{table:italian_regulation}
  \renewcommand{\arraystretch}{1.7} 
  \begin{tabular}{| m{4em} | m{10em} | m{10em} | m{8em} | m{10em} | m{8em} |}
    \hline
    \textbf{Stages} & \textbf{Stay home if sick, \newline Practice good hygiene} & \textbf{Wear facial coverings} & \textbf{Social distancing} & \textbf{Avoid gathering size \newline (Risk   : number)} & \textbf{Locked locations} \\
    \hline
    Stage 0 & False & False & None & None & None\\
    \hline
    Stage 1 & True & False & 0.2 & None & None\\
    \hline
    Stage 2 & True & False & 0.25 & None & School\\
    \hline
    Stage 3 & True & True & 0.6 & Low: 0, High: 0 & School, Hair Salon, Retail Store\\
    \hline
    Stage 4 & True & True & 0.8 & Low: 0, High: 0 & Office, School, Hair Salon, Retail Store\\
    \hline
  \end{tabular}
\end{table*}

\subsection{Sensitivity Analysis}
\label{sec:sensitivity-analisys}

\commentp{Each appendix should start with a pointer to which section of the paper it's relevant to and begin with a description of what is to appear in the appendix. -- FIXED RC}

\begin{figure*}[t]
  \centering
  \begin{subfigure}[b]{0.48\textwidth}
    \includegraphics[width=\textwidth]{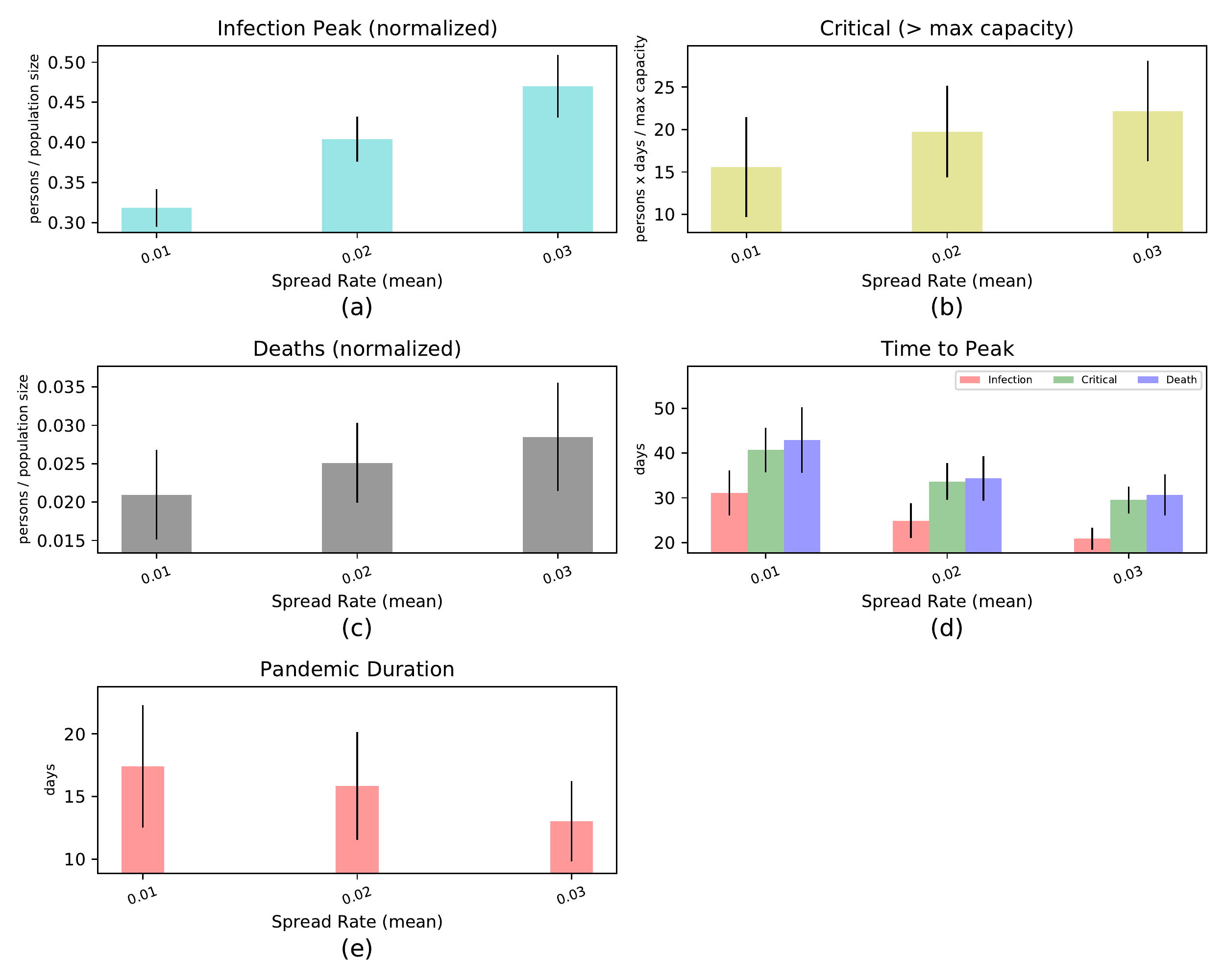}
    \caption{Sensitivity to infection spread rates}
    \label{fig:spread_sensitivity}
  \end{subfigure}
  \begin{subfigure}[b]{0.48\textwidth}
    \includegraphics[width=\textwidth]{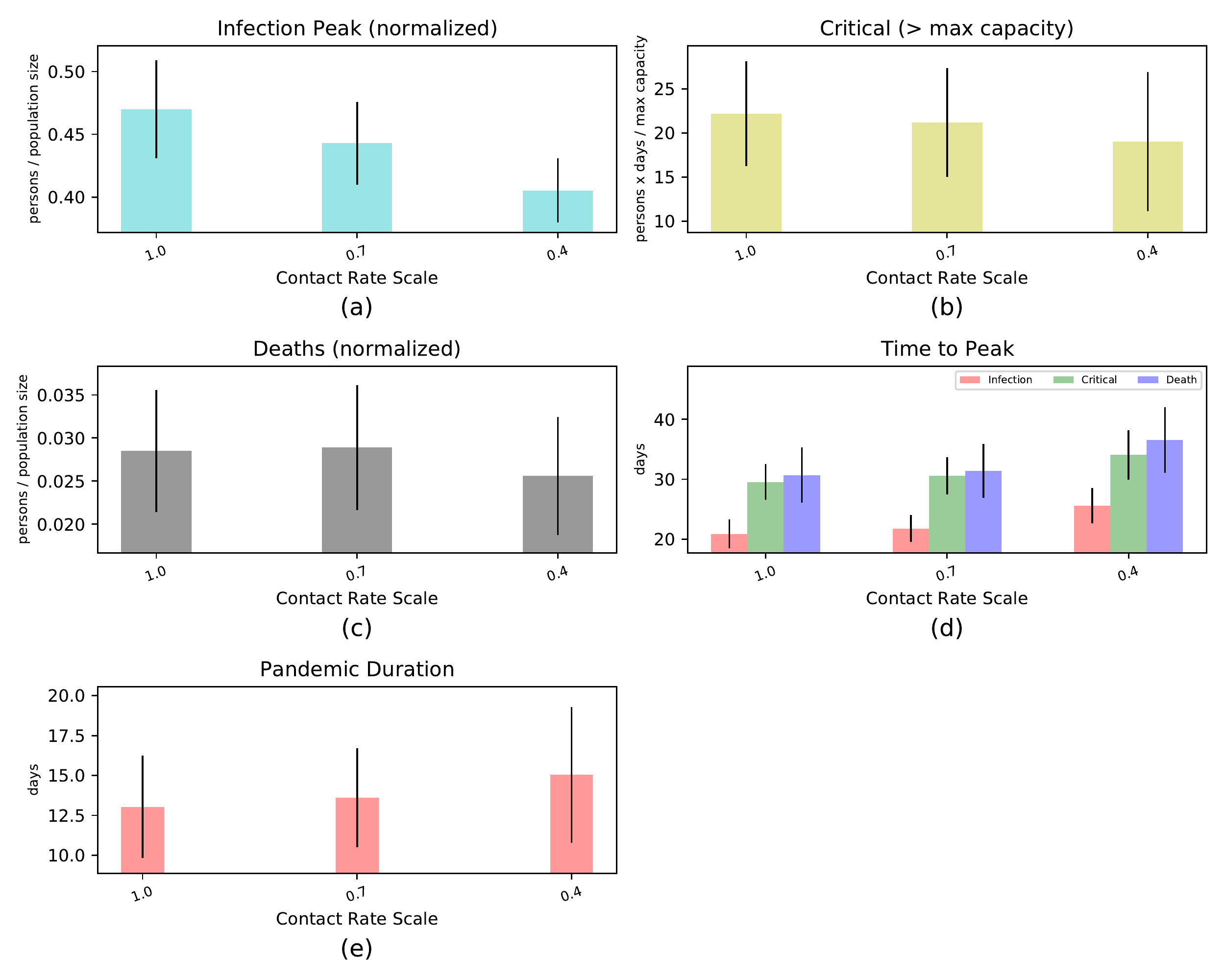}
    \caption{Sensitivity to location's contact rates}
    \label{fig:contact_sensitivity}
  \end{subfigure}
  \caption{Simulator sensitivity to infection spread and contact rates. The
    plots are generated based on 30 different randomly seeded runs of
    the simulator. Mean is shown by a solid line and variance either
    by a shaded region or an error line.}
\end{figure*}

As described in Section \ref{sec:calibration}, we calibrate the simulator to match
the historical data that has been observed around the world. While we describe the calibration process in the next section, here we discuss a preliminary analysis of the sensitivity of \oursim\ to a few of its important parameters needed for the calibration, namely: (1) spread rates for each person, (2) contact rates for each location, (3) social gatherings
size for home parties. Not only this allows us to calibrate \oursim\ but it also enables one to establish that the effects of these parameters are roughly as expected. \commentp{We should explain why we do this analysis - to establish with some confidence that the effects of these parameters is roughly as expected. -- ADDED RC} In more detail, we observe how uniformly
scaling down or up the default value of these parameters gets
reflected in the spread of the pandemic. Then, we use this information
to perform a very simple calibration of the simulator against real
data.

\figurename~\ref{fig:spread_sensitivity}, for example, shows that
increasing individual spread rates results in increased and faster
pandemic peaks. Faster peaks also induce higher numbers of
simultaneous critical cases, that easily pass the hospital capacity
threshold. Consequently, we observe from our plots that spread rates
are directly proportional to number of deaths in \oursim. Note
that this parameter allows us to model super-spreaders on an
individual basis. Thus, increasing spread rates practically means
increasing the number of super-spreaders and easily infecting most of the
simulator's population.

\begin{figure*}[t]
  \centering
    \includegraphics[width=0.9\textwidth]{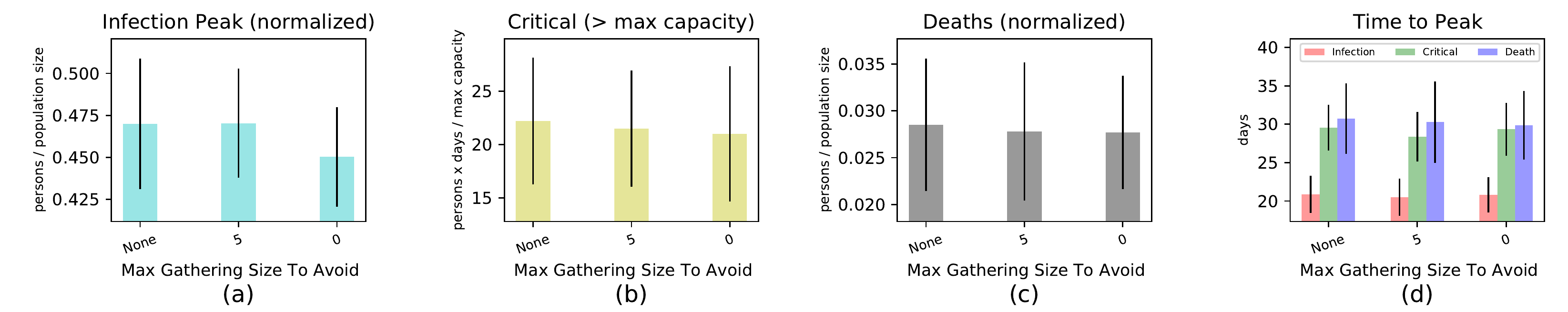}
  \caption{Simulator sensitivity to social gathering sizes through social events. 
  The plots are generated based on 30 different randomly seeded runs of
    the simulator. Mean is shown by a solid line and variance either
    by a shaded region or an error line.}
    \label{fig:social_gatherings_sensitivity}
\end{figure*}

On the contrary, uniformly decreasing contact rates for all locations through a multiplier
has the natural effect of reducing the spread of the pandemic, as well as the number of deaths
(see \figurename~\ref{fig:contact_sensitivity}). This is due to the reduction
of contacts among people in the same location, which is also one the implications of
social distancing. From our plots, it is possible to observe how the relation
between contact rates and number of deaths / critical cases above maximum hospital capacity
is non-linear. This non-linearity arises from the limited degrees of separation of our 
population, which is modeled as a small community where it is easy to be a $2^{\text{nd}}$-degree
contact of an infected person when contact rates are not decreased too much.

Social gathering size is another parameter that we analyze in \oursim. To this end, 
we compare our metrics with maximum size set at: none (no limit), 5, and 0. As shown in
\figurename~\ref{fig:social_gatherings_sensitivity}, also in this case the relation between
the maximum allowed gathering size and deaths / critical cases is non-linear, for the same
reasons expressed above. Moreover, the effect of reducing gathering size is even more limited when compared to contact rate results. This is due to the contact rates remaining unchanged,
reducing social gatherings to yet another location where the virus can be spread. This suggests
that forbidding social gatherings without reducing contact rates or locking other locations
might not be productive.

\subsection{Calibration}
\label{sec:app-calibration}

\commentp{Similarly, we should add a sentence about why calibration is important. Roberto: Done, can you double check it serves the purpose? I took it from related work} For  any  model  to  be  accepted  by  real-world  decision-makers,  they  must  be  provided  with  a  reason  to  trust  that it accurately models the population and spread dynamics in their own community. This trust is typically best instilled via a model calibration process that ensures that the model accurately tracks past data. Based on our parameter sensitivity analysis, we choose to calibrate the 
simulator by directly controlling the mean of the spread rate distribution, while keeping variance 
fixed at $0.01$. In fact, while we do not aim at a precise and complex calibration procedure, this is the only 
parameter among the analyzed ones that has a linear relation to the number of critical cases and deaths,
since the parameters of our infection model are already carefully tuned based on previous work
reported in Table~\ref{table:seir_params}. Specifically, in order to further improve the realism of 
our model, we compare the average time to peak for deaths and tested critical cases in \oursim\ 
against real data from Sweden, as provided by the World Health Organization\footnote{\url{https://covid19.who.int/region/euro/country/se}}. We choose Sweden as our source of data as it
represents the nation where the least restrictions have been applied during the first pandemic
wave and, thus, where the dynamics of the virus is the most ``natural''. In order to match real data
(time-to-peak deaths $\approx 30$ days) we run a coarse grained grid search on the spread rate mean in the range $[0.01, 0.03]$, resulting
in a final parameter choice of $0.03$ (see \figurename~\ref{fig:spread_sensitivity}(d), where the simulator's time-to-peak deaths $\approx 30$ days).\commentp{Can we end this section with some evidence that the simulator is somewhat calibrated?}

\subsection{Testing and Contact Tracing}
\label{sec:app_testingcontact_tracing}

\commentpw{Re-summarize information about the contact tracing and testing results. Original copied text below. - RC FIXED}

\begin{figure*}
  \centering
  \includegraphics[width=0.9\textwidth]{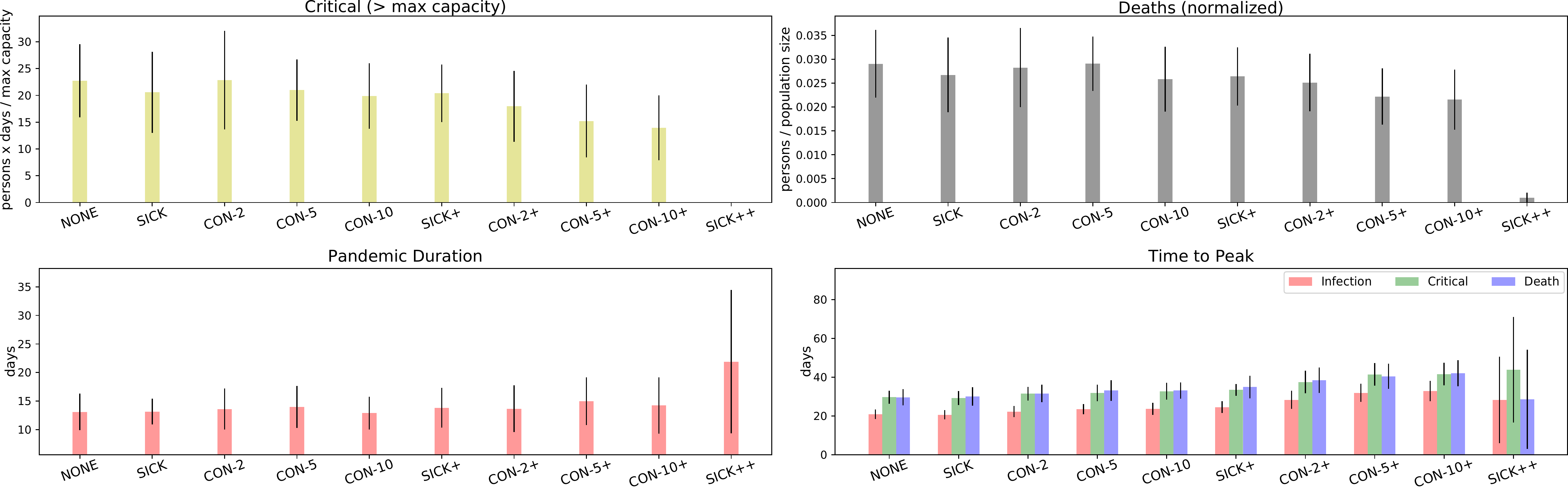}
  \caption{Comparison of various combinations of testing and contact tracing.}
  \label{fig:contact_tracing}
\end{figure*}

As described in Section \ref{sec:testing-contact-tracing-res}, to validate \oursim's ability to model testing and contact tracing, we compare here several strategies with different testing
rates and contact horizons. Specifically, we consider daily testing rates of \{0.02, 0.3 (denoted with a + symbol in our plots), and 1.0 (++)\} (where 1.0 represents the extreme case of everyone being tested every day) and contact tracing histories of \{0, 2, 5, or 10\} days. Note that, in our plots, both NONE and SICK use 0 contact tracing history (the second with self-isolation at symptom onset), while CON-$N$ uses an $N$ length history. The full specification of each parameter combination as well as the mapping to the label in \figurename~\ref{fig:contact_tracing} is shown in Table~\ref{table:contact_tracing_experiment}. 
For each condition, we ran the experiments with the same 30 random seeds. 

Not surprisingly, contact tracing is most beneficial with higher testing rates and longer contact histories because more testing finds more infected people and the contact tracing is able to encourage more of that person's contacts to stay home. Of course, the best strategy is to test every person every day and quarantine anyone who tests positive (SICK++). Unfortunately, this strategy is impractical except in the most isolated communities. It is interesting to note that this strategy has the highest variance in the time to peak and pandemic duration plots due to the fact that, in this scenario, the virus either stops immediately, or it spreads very slowly.

\begin{table*}[hb!]
\footnotesize
  \caption{Testing and contact tracing policies.}
  \label{table:contact_tracing_experiment}
  \renewcommand{\arraystretch}{1.7} 
  \begin{tabular}{| c | c | c | c | c |}
    \hline
    \textbf{Name} & \textbf{Contact tracing history (days)} & \textbf{Random testing rate (daily)} & \textbf{Stay home if sick} & \textbf{Stay home if positive contact} \\
    \hline
    \textbf{NONE} & 0 & 0.02 & NO & NO \\
    \hline
    \textbf{SICK} & 0 & 0.02 & YES & NO \\
    \hline
    \textbf{CON-2} & 2 & 0.02 & YES & YES \\
    \hline
    \textbf{CON-5} & 5 & 0.02 & YES & YES  \\
    \hline
    \textbf{CON-10} & 10 & 0.02 & YES & YES  \\
    \hline
    \textbf{SICK+} & 0 & 0.3 & YES & NO \\
    \hline
    \textbf{CON-2+} & 2 & 0.3 & YES & YES  \\
    \hline
    \textbf{CON-5+} & 5 & 0.3 & YES & YES  \\
    \hline
    \textbf{CON-10+} & 10 & 0.3 & YES & YES  \\
    \hline
    \textbf{SICK++} & 0 & 1 & YES & NO \\
    \hline
    \end{tabular}
\end{table*}

\begin{figure*}[t]
  \centering
  \includegraphics[width=0.7\textwidth]{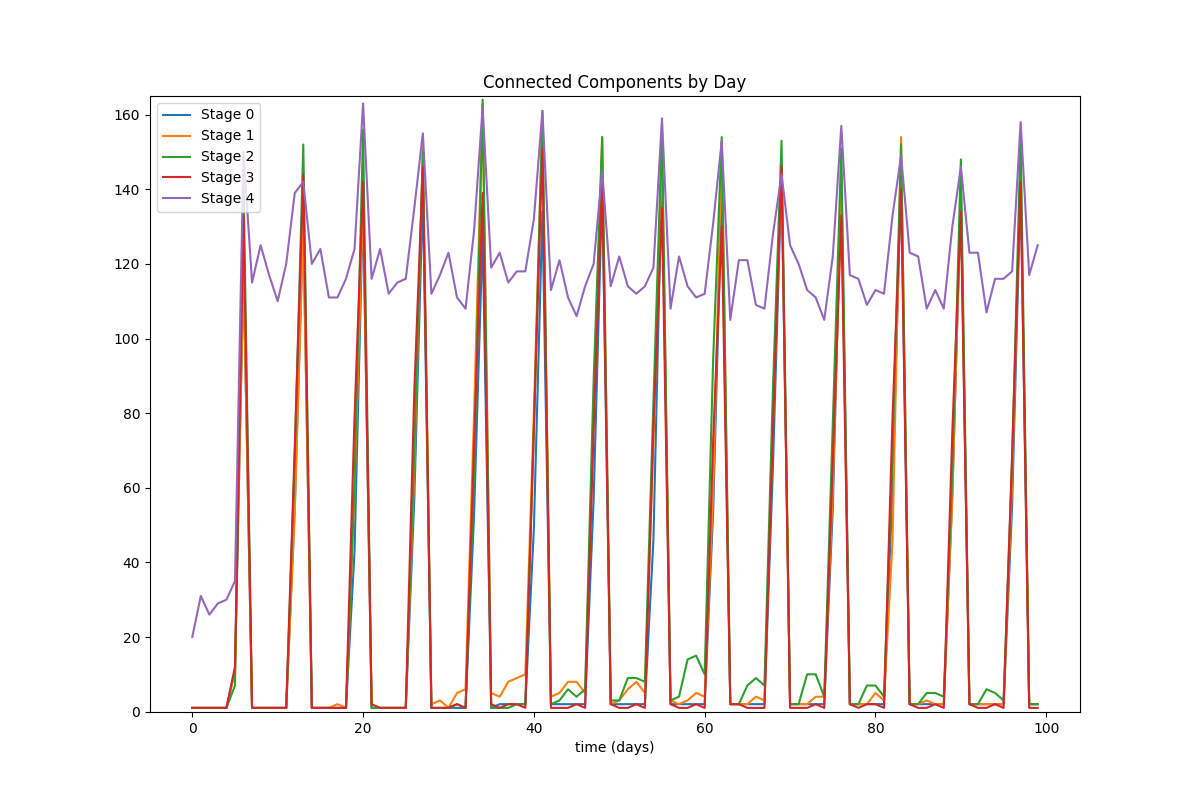}
  \caption{Graph connectivity over 5 different runs, each at a different stage.}
  \label{fig:connectivity}
\end{figure*}

\subsection{Graph Connectivity}
\label{sec:connectivity}

One of the ways in which the effects of governmental restrictions can be 
understood is that they influence the connectivity of the contact graph 
among members of the population. For example, if the graph is fully 
connected, the average degree of separation between individuals may be 
increased by increasing restrictions, which presumably would then slow 
the spread of the disease. 
 
To investigate the extent to which the restrictions we model influence 
the graph connectivity, we collected interaction data from runs corresponding to those in \figurename~\ref{fig:staged_experiment}. From these, we generated a graph of all interactions between the people in the 
simulator on each day of the simulation.
 
Our findings indicate that, in fact, the graph is often not connected.  
We therefore analyze the number of connected components in the 
interaction graphs during 5 separate runs, each at a different stage.
The results are plotted in \figurename~\ref{fig:connectivity}. 
 
As is apparent in the graph, on weekends (the periodic peaks in the 
graph), the interaction graph has many many separate components, 
indicating a greater degree of separation between people. Similarly 
during Stage 4 restrictions, the number of connected components is quite 
high, suggesting that lockdowns can be very effective in slowing the 
spread of the pandemic. Somewhat surprisingly, on weekdays at other 
stages, the number of connected components is relatively low, with 
relatively small differences between the stages. 
 
An interesting direction for future work is to do a more in-depth graph 
connectivity analysis, including for interaction graphs that span 
multiple days. 
\commentpw{It is hard to draw much from the figure. Stage 4 creates a lot of small disconnected subgraphs throughout the simulation. The others are essentially fully connected during the week and break into subgraphs on the weekends.}

\subsection{Simulation Time}
In Section~\ref{sec:experiments}, we mention that \oursim\ can easily handle larger experiments at the cost of greater time and computation. In Table~\ref{table:execution_times}, we report simulation times for 1k, 2k, 4k and 10k population environments. For 1k, we also report the training time at which our reinforcement learning algorithm converges.

\begin{table*}[]
    \centering
    \caption{Simulation time for different population sizes. The simulator was run on a single core Intel i7-7700K CPU $@$ 4.2GHz with 32GB of RAM.}
    \begin{tabular}{|c|c|}
    \hline
      \textbf{Population size}  &  \textbf{Simulation Time} \\
      \hline
      1k & 25.4 msecs/sim-step (our RL training took about 4 hours to converge)\\
      \hline
      2k & 57.9 msecs/sim-step\\
      \hline
      4k & 138.5 msecs/sim-step \\
      \hline
      10k & 500 msecs/sim-step\\
      \hline

    \end{tabular}
    \label{table:execution_times}
\end{table*}

\end{document}